\documentclass{article}

\usepackage{microtype}
\usepackage{graphicx}
\usepackage{subfigure}
\usepackage{caption}
\usepackage{booktabs} 
\usepackage{multirow}

\usepackage{amsmath}
\usepackage{amsfonts}

\usepackage{hyperref}

\usepackage[accepted]{icml2019}

\icmltitlerunning{Compressing GANs using Knowledge Distillation}

\begin{document}

\twocolumn[
\icmltitle{Compressing GANs using Knowledge Distillation}



\icmlsetsymbol{equal}{*}

\begin{icmlauthorlist}
\icmlauthor{Angeline Aguinaldo*}{umd,jhuapl}
\icmlauthor{Ping-Yeh Chiang*}{umd}
\icmlauthor{Alex Gain*}{jhu}
\icmlauthor{Ameya Patil*}{umd}
\icmlauthor{Kolten Pearson*}{umd}
\icmlauthor{Soheil Feizi}{umd}
\end{icmlauthorlist}

\icmlaffiliation{umd}{Department of Computer Science, University of Maryland, College Park}
\icmlaffiliation{jhu}{Department of Computer Science, Johns Hopkins University}
\icmlaffiliation{jhuapl}{Johns Hopkins University Applied Physics Laboratory}

\icmlcorrespondingauthor{Angeline Aguinaldo}{aaguinal@cs.umd.edu}
\icmlcorrespondingauthor{Ping-Yeh Chiang}{pchiang@cs.umd.edu}
\icmlcorrespondingauthor{Alex Gain}{again1@jhu.edu}
\icmlcorrespondingauthor{Kolten Pearson}{kolten@cs.umd.edu}
\icmlcorrespondingauthor{Ameya Patil}{ameyap@cs.umd.edu}
\icmlcorrespondingauthor{Soheil Feizi}{sfeizi@cs.umd.edu}

\icmlkeywords{Machine Learning, GANs, knowledge distillation, compression, student-teacher, ICML}

\vskip 0.3in
]



\printAffiliationsAndNotice{\icmlEqualContribution} 

\begin{abstract}

Generative Adversarial Networks (GANs) have been used in several machine learning tasks such as domain transfer, super resolution, and synthetic data generation. State-of-the-art GANs often use tens of millions of parameters, making them expensive to deploy for applications in low SWAP (size, weight, and power) hardware, such as mobile devices, and for applications with real time capabilities. There has been no work found to reduce the number of parameters used in GANs. Therefore, we propose a method to compress GANs using knowledge distillation techniques, in which a smaller ``student'' GAN learns to mimic a larger ``teacher'' GAN. We show that the distillation methods used on MNIST, CIFAR-10, and Celeb-A datasets can compress teacher GANs at ratios of 1669:1, 58:1, and 87:1, respectively, while retaining the quality of the generated image. From our experiments, we observe a qualitative limit for GAN's compression. Moreover, we observe that, with a fixed parameter budget, compressed GANs outperform GANs trained using standard training methods. We conjecture that this is partially owing to the optimization landscape of over-parameterized GANs which allows efficient training using alternating gradient descent. Thus, training an over-parameterized GAN followed by our proposed compression scheme provides a high quality generative model with a small number of parameters. 
\end{abstract}

\section{Introduction}
Generative Adversarial Networks (GANs) were first introduced in 2014 as a generative model that attempts to capture the underlying distribution of complex real world data sets \cite{goodfellow2014generative}. 
GANs are applied to many real world applications including domain transfer, super resolution, and generation of novel celebrity faces \cite{karras2018progressive} \cite{CycleGAN2017} \cite{ledig2017photo}. 
The first GAN implemented by Ian Goodfellow, used no more than 100,000 parameters, otherwise referred to as hidden units or neurons \cite{goodfellow2014generative}. In one of NVIDIA's most recent publications, the company's network used over 23 million parameters to generate realistic celebrity faces with high resolution \cite{karras2018progressive}. Most recently, Google's BigGAN has over 158 million parameters used to generate photo-realistic still life imagery \cite{Brock2018}. We anticipate that the size of GANs will continue to increase, as their applicability continues to grow. If we want to make GANs practical for low SWaP (size, weight, and power) hardware, such as mobile devices, and for applications with real time capabilities then being able to compress models becomes an important issue to overcome. 

There has been recent progress in the area of neural network compression \cite{liu2019model} \cite{ba2013deep} \cite{urban2017dothey} and various compression techniques have been exercised to solve this problem \cite{Belagiannis2018} \cite{zheng2017cond} \cite{yim2017kd} \cite{kim2017tranfer}. The popular compression techniques can be mostly categorized into five schemes: quantization \cite{Goldstein2017quantization}, pruning and sharing weights, low rank factorization, compact convolutional filter, and knowledge distillation \cite{cheng2017survey}. 
These techniques are able to reduce the number of parameters by up to 90\% to 95\% while retaining performance \cite{cheng2017survey}. Some of these compression techniques also accompany corollary benefits. In the case of knowledge distillation, the compressed model may be able to generalize better in addition to being significantly smaller \cite{hinton2015distilling}.

Despite the aforementioned work, to our knowledge, there is no result involving the compression of GANs. This leads to the work described in this paper, where we show adaptations that must be made to the knowledge distillation paradigm in order to achieve optimal compression of networks in the generative setting. Serving as motivation for these adaptations is the idea that large over-parameterized networks have nicer loss landscapes than smaller ones, and are thus able to learn better quality mappings, regardless of whether an approximately equivalent mapping exists for smaller networks. We experimentally validate these methods on several datasets and via a number of objective measurements. Lastly, we discuss the limit of compression in the GAN setting and how it appears in empirical results.


\begin{figure*}[!t]
\vskip 0.2in
\begin{center}
\centerline{\includegraphics[clip, trim=0cm 1.5cm 0cm 0cm,
width=\textwidth]{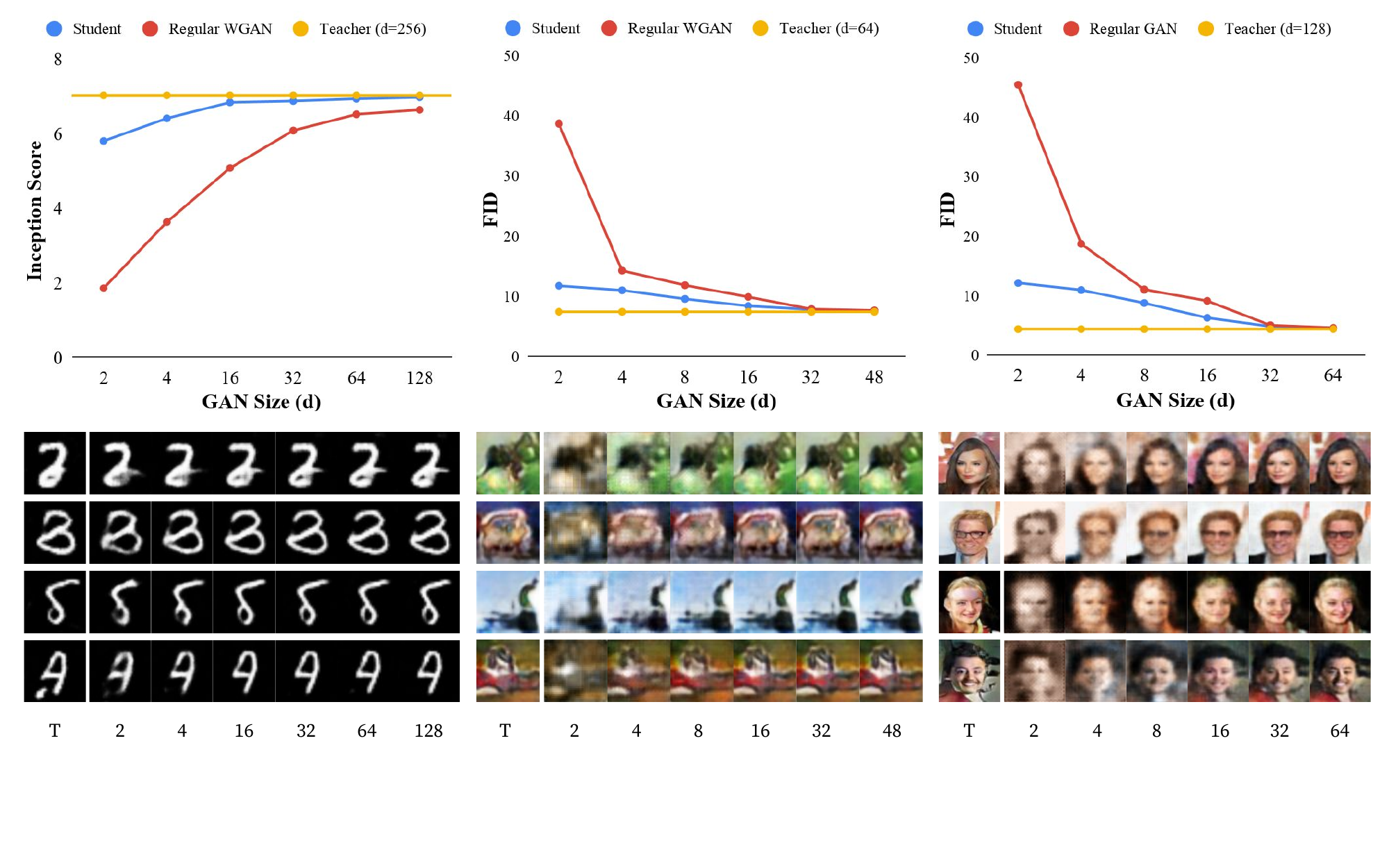}}
\caption{The top row shows Inception Score and Frechet Inception Distance comparison for various sizes of GANs, where GAN size is determined by the depth scale factor, $d$, and 3 different datasets - MNIST, CIFAR-10 and Celeb-A (left to right). The students are trained using the MSE loss training scheme \ref{mse_loss}. A high Inception Score is good and a low Frechet Inception Distance is good. The bottom row shows samples of generated images from the students and teacher GANs as a result of corresponding compression.}
\label{fig:combined_quality_metrics_plot}
\end{center}
\vskip -0.2in
\end{figure*}

\section{Background}

\subsection{Generative Adversarial Networks}
GAN was first proposed as a two player min-max optimization problem between a discriminator $f_w(.)$ and a generator $g_\theta(.)$ as in (\ref{eq:1}) \cite{goodfellow2014generative}. The generator is tasked with generating realistic examples that fool the discriminator while the discriminator learns how to differentiate between the real and the generated samples.
\begin{equation}
\begin{split}
\label{eq:1}
    \min_{\theta \in \Theta} \max_{w \in W} \mathbb{E}_{\mathbf{x}\sim p_{data}}[\log(f_w(\mathbf{x})] \\
    + \mathbb{E}_{\mathbf{z}\sim p_z}[\log(1 - f_w(g_\theta(\mathbf{z})))]
\end{split}
\end{equation}
The optimization in (\ref{eq:1}) has a global minimum and the system converges when \(p_g = p_{data}\) at which point, $f_w(.)$ cannot classify a sample as being generated from \(p_g\) or from \(p_{data}\). Further, the optimal solution to (\ref{eq:1}) corresponds to minimizing the Jensen-Shannon (JS) divergence between the two distributions \(p_{data}\) and \(p_g\) \cite{goodfellow2014generative}.
However, training of GANs is often unstable because JS divergence is not well defined when \(p_g\) and \(p_{data}\) do not have the same support \cite{arjovsky2017wasserstein}. To solve the problem with using JS divergence, WGAN minimizes the Wasserstein's distance between \(p_g\) and \(p_{data}\)  in place of the JS divergence\cite{arjovsky2017wasserstein}, which is well defined even when \(p_g\) and \(p_{data}\) have disjoint support. Specifically, WGAN attempts to solve the optimization problem in (\ref{eq:2}), where $f_w(.)$ is a Lipschitz bounded function. \cite{arjovsky2017wasserstein}.
\begin{equation}
    \min_{\theta \in \Theta} \max_{w\in W} \mathbb{E}_{\mathbf{x}\sim p_{data}}[f_w(\mathbf{x})] - \mathbb{E}_{\mathbf{z}\sim p_z}[f_w(g_{\theta}(\mathbf{z}))]
\label{eq:2}
\end{equation}
WGAN was used in place of regular GAN for most of our experiments due to its favorable characteristics. However, empirically, we noticed that WGAN does not work as well for the Celeb-A dataset, so we reverted to using regular GAN for all Celeb-A related experiments.
\subsection{Knowledge Distillation}
\label{sec:KD}
Knowledge distillation refers to the technique of transferring the knowledge learned, from an ensemble of networks to a single network, or from a network with high number of parameters, to a network with relatively low number of parameters. We refer to the bigger network as the \emph{teacher} network and the smaller network as the \emph{student} network. 

A student can learn to match any activation layer in the teacher network. Learning parameters from the final layer, called hard targets, lends itself to shorter training time but increased chance of over-fitting. The inputs to the softmax layer (logits) of the teacher network, referred to as soft targets, on the other hand, have more descriptive information about the samples and give better generalization characteristics to the student network \cite{hinton2015distilling}, which makes training on soft targets more beneficial. 

\subsection{Over-parameterization of Networks}
An over-parameterized network is described as one whose number of hidden units is polynomially large relative to the number of training samples \cite{allen2018convergence}. It has been shown that training a significantly over-parameterized GAN yields dramatically better results than those generated from a smaller network \cite{Brock2018}. This may be explained by a finding that showed that the over-parameterization of neural networks creates optimized loss functions with many good minima spread throughout the entire loss landscape allowing for efficient training with alternating gradient descent \cite{allen2018convergence} \cite{allen2018learning}. This theory was bolstered by recent empirical studies of loss functions using visualization methods~\cite{li2018visualizing}. Therefore, it is necessary that a bigger network learn these mappings in a hyper-parameterized space before it can be distilled to a simpler model. Likewise, there has been empirical evidence that knowledge distillation, or model compression, is successful \cite{hinton2015distilling} \cite{bucilua2006model} \cite{yim2017kd}. This success may be attributed to the aforementioned phenomena. Although training a teacher network might require a higher number of parameters, a reduced number of parameters is sufficient to describe the model with high fidelity.

\section{Methods}


The teacher (large, over-parameterized network) and student (small, few parameter network) GANs used either the original DCGAN architecture or a slightly modified DCGAN architecture \cite{radford2015unsupervised}, more closely resembling the WGAN \cite{arjovsky2017wasserstein}, referenced as the W-DCGAN. 

The number of parameters in our networks is controlled by the depth scale factor, referenced throughout the paper as $d$. The overall number of parameters increases approximately linearly to $d^2$.




\subsection{Selection of Teacher Network}
\begin{figure*}[!h]
\vskip 0.2in
\begin{center}
\centerline{\includegraphics[clip, trim = 0.5cm 4.5cm 1cm 2.5cm, width=\textwidth]{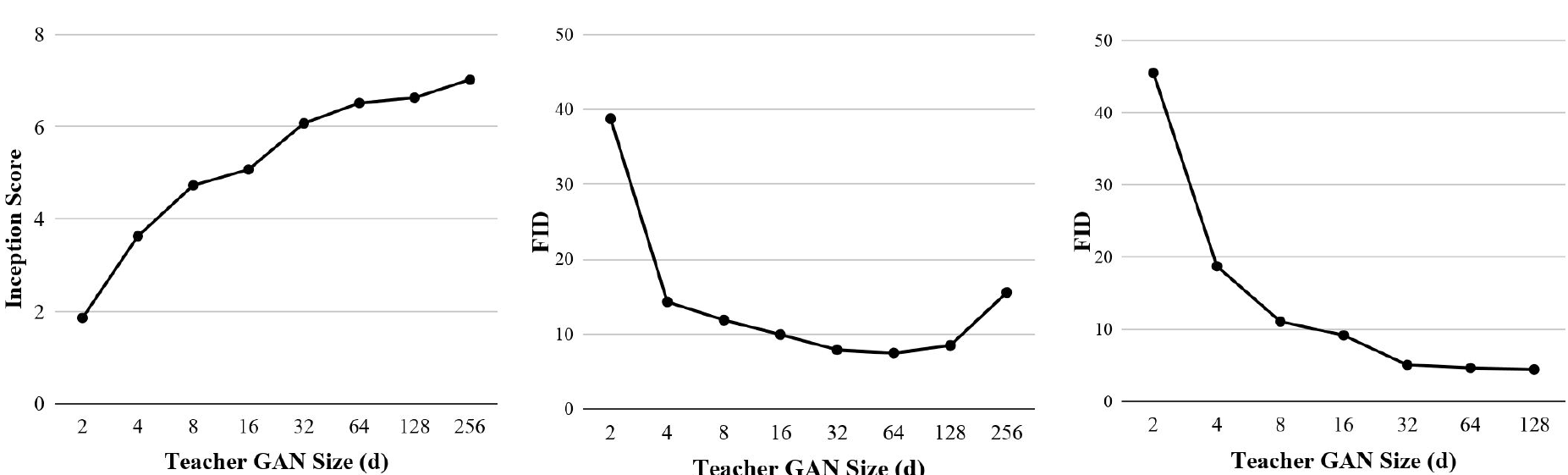}}
\caption{The Inception Score and Frechet Inception Distance was used to evaluate the best teacher GAN, parameterized by the depth scale factor $d$. A high Inception Score is good and a low Frechet Inception Distance is good. From these results, we selected a teacher GAN size of $d=256$ for MNIST, $d=64$ for CIFAR-10 and $d=128$ for Celeb-A (left to right).}
\label{fig:teacher_gan_selection}
\end{center}
\vskip -0.2in
\end{figure*}

We first trained W-DCGANs of various sizes until convergence, and then selected the best performing model to be the teacher network. This ensures that the teacher network has converged approximately to an optimal solution. Because there currently does not exist an exact measure of visual quality, we use Inception Score and Frechet Inception Distance as proxies for performance. Figure \ref{fig:teacher_gan_selection} illustrates the Inception Score performance and Frechet Inception Distance (FID) performance respectively, to be discussed in Section \ref{sec:eval_metrics}, with respect to layer depth, $d$.

\subsection{Training of Student Networks}
We train several student networks with smaller capacities than the teacher network using two training schemes. Results for all three datasets were produced using the MSE Loss training scheme. Due to the complexity of the Celeb-A dataset, the joint loss training scheme was designed to combat observed blur artifacts of compression. Both training schemes are described below. For both functions, we monitored the convergence of the student network based on the generated outputs and the loss trajectory. 
\begin{figure}[!ht]
    \centering
    \vskip 0.2in
    \centerline{\includegraphics[clip, trim=0cm 8cm 1.5cm 0cm, width=\columnwidth]{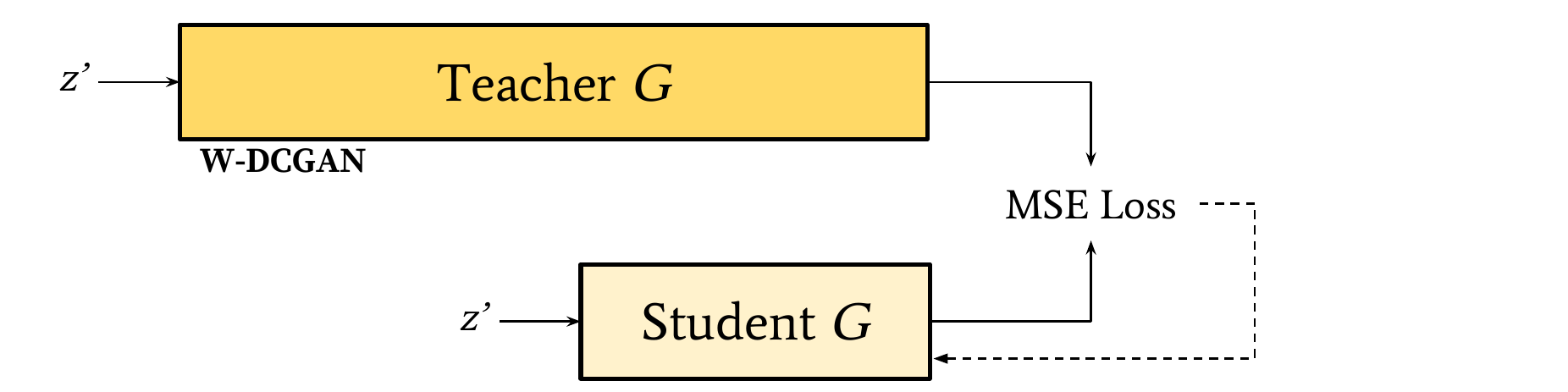}}
    \caption{Student-teacher training framework with mean squared error (MSE) loss for student training. The teacher generator was trained using DCGAN framework \cite{radford2015unsupervised} including WGAN modifications \cite{arjovsky2017wasserstein}. A mathematical analogy is shown in \eqref{eq:mse}.}
    \label{fig:st_mse_loss}
    \vskip -0.2in
\end{figure}

\label{mse_loss}
\emph{Mean Squared Error (MSE) loss.} This method uses the MSE as the student training loss function using a pre-trained teacher W-DCGAN. A schematic of the training framework is illustrated in Figure \ref{fig:st_mse_loss}. The MSE loss minimizes the pixel-level error between the images generated from the student and the teacher. Specifically, we train the student by solving the following optimization problem:
\begin{equation}
    \min_{\theta} \mathbb{E}_{\mathbf{z}\sim N(0,I)}
    \Big[
        \Big\Vert g_{teacher}(\mathbf{z})-g_{\theta}(\mathbf{z})\Big\Vert^{2}
    \Big]
\label{eq:mse}
\end{equation}
\emph{Joint loss.} The generated images tend to be slightly blurry when using the MSE loss, especially for the Celeb-A dataset. To combat the blurriness, we propose a joint loss function that supervises regular GAN training with MSE loss. Specifically, the joint loss train the student by solving the following optimization problem:
\begin{figure}[!t]
    \centering
    \vskip 0.2in
    \centerline{\includegraphics[clip, trim=0cm 5cm 0cm 0cm, width=\columnwidth]{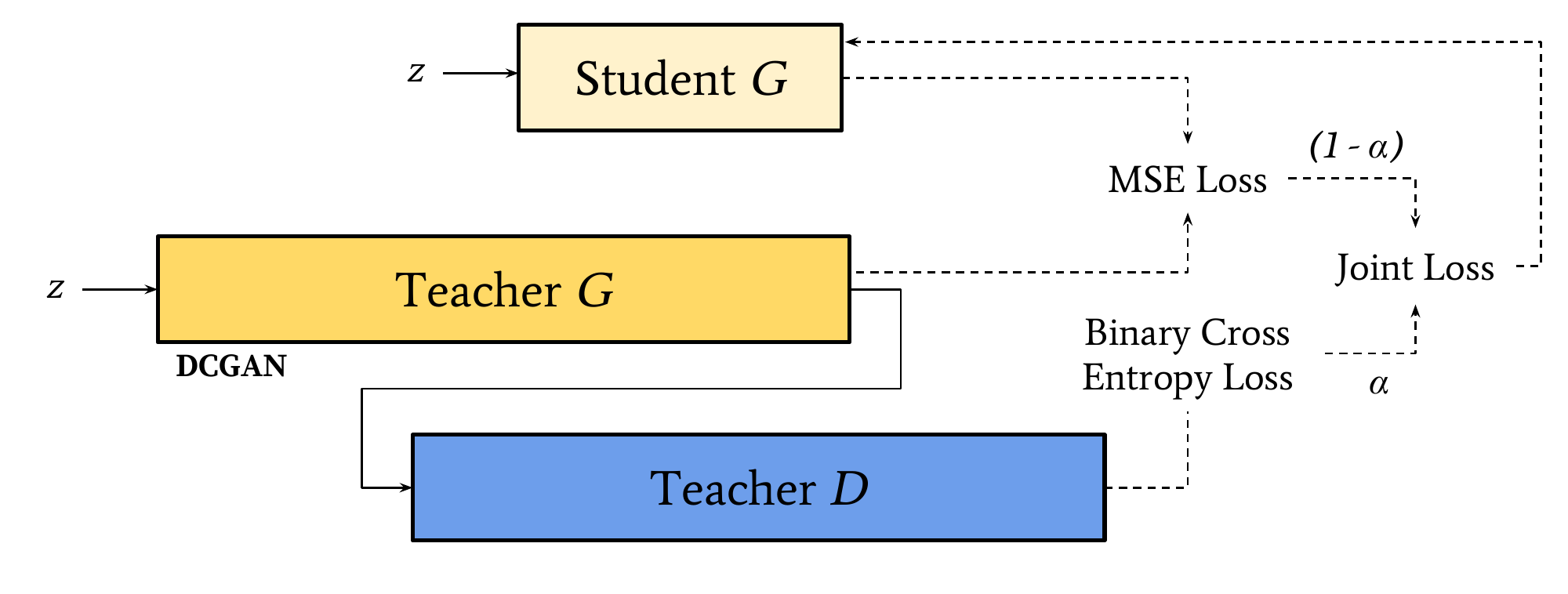}}
    \caption{Student-teacher training framework with joint loss for student training. The teacher generator was trained using DCGAN framework \cite{radford2015unsupervised}. A mathematical analogy is shown \eqref{eq:joint_training}.}
    \label{fig:st_joint_loss}
    \vskip -0.2in
\end{figure}
\begin{equation}
\centering
\begin{split}
    \min_{\theta \in \Theta} \max_{w \in W} \mathbb{E}_{\mathbf{x}\sim p_{data}}[\log(f_w(\mathbf{x})] \\
    + \mathbb{E}_{\mathbf{z}\sim p_z}[\alpha\log(1 - f_w(g_\theta(\mathbf{z}))) \\
    +(1-\alpha)
        \Big\Vert g_{teacher}(\mathbf{z})-g_{\theta}(\mathbf{z})\Big\Vert^{2}
    ]
\end{split}
\label{eq:joint_training}
\end{equation}
The $\alpha$ parameter controls the weight between the MSE loss and the regular GAN training. A schematic of the training framework is illustrated in Figure \ref{fig:st_joint_loss}. 
\section{Analysis}
\label{sec:eval_metrics}
In the case of classification networks, the performance can be measured by the classification accuracy. Unlike classification networks, GANs do not have an explicit measure for performance. The performance of GANs could be naively measured by human judgment of visual quality \cite{goodfellow2014generative}. For example, one could collect scores (1 to 10) of visual quality from various subjects and average the scores to understand the performance of GANs. However, the method is very expensive. The score could also vary significantly based on the design of the interface used to collect the data \cite{goodfellow2014generative}. To evaluate the performance of GANs more systematically, the field has developed several quantitative metrics. Some of the popular metrics are Inception Score and Frechet Inception Distance (FID). Additionally, we used Variance of Laplacian to evaluate the blurring artifacts inherent to compressing GANs trained on complex datasets.
\begin{figure*}[!ht]
\vskip 0.2in
\begin{center}
\centerline{\includegraphics[width=\textwidth]{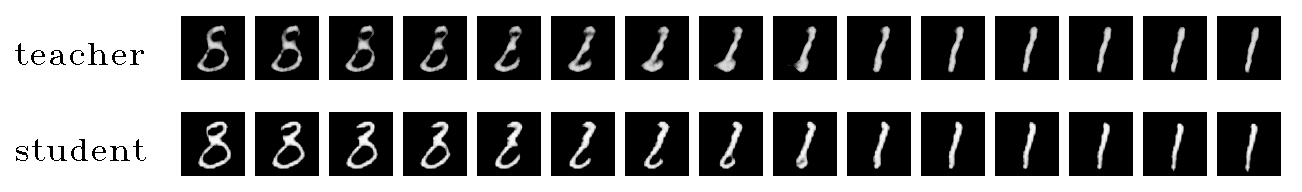}}
\caption{A comparison of the student and the teacher on inputs interpolated between a representation of "8" and a representation of "1". 
}
\label{fig:mnist_interpolation}
\end{center}
\vskip -0.2in
\end{figure*}

\subsection{Inception Score (IS)}
There are two important things that we would like to see in images generated from good GANs. First, we would like it to generate diverse images. We would like $p(\mathbf{y})$ to be relatively equal across different classes \cite{goodfellow2014generative}. Secondly, given a generated image, we would like to be confident of the class in which the image belongs. Given a generated image x, we would like $p(\mathbf{y}|\mathbf{x})$ to be very concentrated in a particular class \cite{goodfellow2014generative}. To take both of the desired qualities into account,  the cross entropy, $H(\cdot,\cdot)$, between $p(\mathbf{y})$ and $p(\mathbf{y}|\mathbf{x})$ can be taken, otherwise known as the Inception Score. 
\begin{equation}
\textit{IS} = H(p(\mathbf{y}), p(\mathbf{y}|\mathbf{x}))
\end{equation}
If $p(\mathbf{y})$ is similar across classes and $p(\mathbf{y}|\mathbf{x})$ is very concentrated in a particular class, then the cross entropy between the two distributions will be high. Consequently, the Inception Score will be high.

The Inception Scores makes a few assumptions. First, it assumes that the image can be classified yielding $p(\mathbf{y})$ and $p(\mathbf{y}|\mathbf{x})$, but not all images can be classified. For example, in our experiments with the Celeb-A dataset, we could not use Inception Score because the data set does not have labels associated with them. Second, the Inception Score is sensitive to the weights of the classification network. The Inception Score assumes a network architecture designed for classification, and the Inception Score could change drastically with different networks. Further, the calculation of Inception Score assumes that the soft-max layer of a neural network is equivalent to the probability distribution. It can be said that soft-max layer of a neural network is not necessarily the probability distribution despite summing to unity.

Finally, the Inception Score is not able to detect memorization of examples. For example, if a GAN remembers exactly one image from each class, then the Inception Score will be very high as $p(\mathbf{y})$ will be exactly the same across all classes and $p(\mathbf{y}|\mathbf{x}))$ will be very concentrated.

Despite the aforementioned, it is one of the most commonly used metrics for GAN evaluation. It is also found to correlate well with human judgment of image qualities \cite{goodfellow2014generative}. Therefore, we used Inception Score to select the best MNIST teacher model and to evaluate the MNIST student models.

\subsection{Frechet Inception Distance (FID)}

\label{sec:FID}
To improve upon the Inception Score, Frechet Inception Distance was introduced to identify GANs that simply memorized a few images from each class \cite{heusel2017gans}. The Frechet Inception Distance assumes that when differing images are fed through the same network, their corresponding values from the same activation layer will have different distributions. If the activation distributions of the generated images and the real images differ greatly, then it is likely that the generated images look significantly different from the real images and vice versa. Formally, Frechet Inception Distance measures the difference of the activation distributions with Frechet distance \cite{heusel2017gans}:
\begin{equation}
\textit{FID} = ||\mu_r -\mu_g||^2+Tr[\Sigma_r + \Sigma_g - 2 (\Sigma_r\Sigma_g)^{1/2} ]
\end{equation}
Empirically, it has been shown that Frechet Inception Distance almost always increases monotonically as you increase the distortion to a real images (such that they look less like real images), regardless of the type of distortion applied \cite{heusel2017gans}. Additionally, it is robust in detecting mode collapse in GANs \cite{luvcic2017gans}.

Though Frechet Inception Distance still suffers from similar downsides as Inception Score, its ability to identify mode collapse makes it more robust compared to Inception Score. It is used in our experiments to select the best teacher GAN and to evaluate the performance of the student GANs for CIFAR-10 and Celeb-A datasets.

\subsection{Variance of Laplacian (VoL)}
The variance of laplacian of an image gives a measure of the sharpness of the image \cite{pacheco2000diatom}. The Laplacian filter, when applied on an image, gives the second order derivative of the discrete image function. It thus highlights the regions of an image containing rapid intensity changes - the edges. A sharper image will have more well defined edges than a blurred image. The Variance of Laplacian metric quantifies the amount of edges in an image. Higher VoL corresponds to sharper images. We use this metric to compare the outputs of the student generator trained using the MSE loss versus those trained using the joint loss.

\section{Results}

\begin{figure*}[!hp]
    \centering
    \begin{subfigure}
        \centering
        \includegraphics[width=\textwidth]{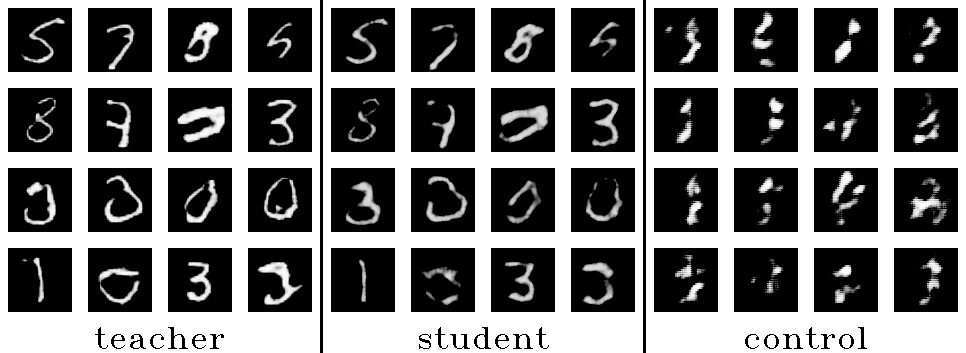}
        \caption{Comparison between results from teacher GAN $d=256$, student GAN $d=2$, and a regularly trained GAN of depth $d=2$ (control) on the MNIST dataset using the MSE loss training scheme as described in Figure \ref{fig:st_mse_loss}.}
        \label{fig:output_comparison_mnist}
    \end{subfigure}
    
    \begin{subfigure}
        \centering
        \includegraphics[width=\textwidth]{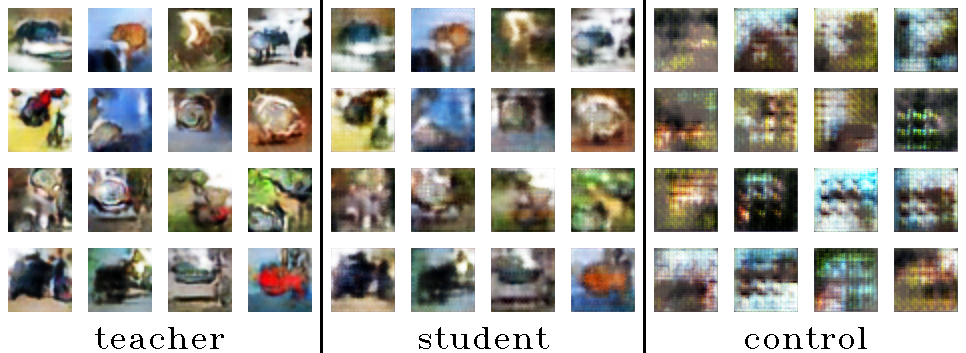}
        \caption{Comparison between results from teacher GAN $d=64$, student GAN $d=4$, and a regularly trained GAN of depth $d=4$ (control) on the CIFAR-10 dataset using the MSE loss training scheme as described in Figure \ref{fig:st_mse_loss}.}
        \label{fig:output_comparison_cifar}
    \end{subfigure}
    
    \begin{subfigure}
        \centering
        \includegraphics[width=\textwidth]{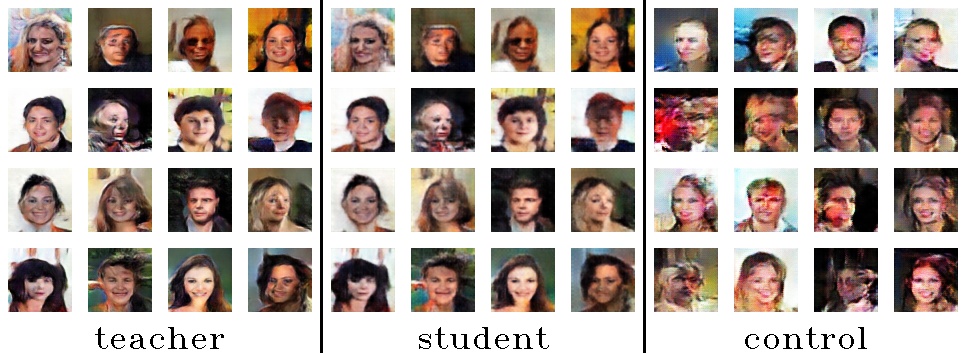}
        \caption{Comparison between results from teacher GAN $d=128$, student GAN $d=16$, and a regularly trained GAN of depth $d=16$ (control) on the Celeb-A dataset using the MSE loss training scheme as described in Figure \ref{fig:st_mse_loss}.}
        \label{fig:output_comparison_celeba}
    \end{subfigure}
    
\label{fig:output_comparison}
\end{figure*}

\subsection{MSE Loss Training}

\emph{Quantitative Results}. We compare the performance of the compressed GANs with the teacher WGAN and the regular WGANs of the same sizes. Ideally, the compressed GAN will perform close to the teacher WGAN and better than the regular WGANs of the same sizes. Again, because there currently does not exist an exact measure of visual quality, we use Inception Score and Frechet Inception Distance as proxies for performance.

From Figure \ref{fig:combined_quality_metrics_plot} and Table \ref{table:compress_ratio}, we can see that the student GANs consistently outperform the regular GAN across all compression level on the MNIST data set. Also, the student GANs perform comparable to the teacher GAN which has significantly larger capacity. In the most extreme case for MNIST, we were able to compress the student model by 1,669 times while retaining 83\% of the teacher's Inception Score.


Similarly, we see that the student GANs consistently outperform the regular GANs of the same sizes, on the CIFAR-10 and Celeb-A dataset. However, because the CIFAR-10 and Celeb-A datasets have images with more complex features than the MNIST dataset, we were unable to achieve the same magnitude of compression as that of the MNIST dataset. 

We found that the compression strategy is extremely robust to hyperparameter tuning, datasets, and evaluation metrics. The compressed student models consistently outperform W-DCGANs of similar size with respect to all changes in our setups.

\begin{table*}
    \caption{Compression Ratios and Image Quality Metrics for MNIST, CIFAR-10, Celeb-A. The respective metrics (\emph{IS}, Inception Score and \emph{FID}, Frechet Inception Distance) are shown for both the student GAN (Stu.) compared to a regularly trained GAN (Reg.) of corresponding size. For MNIST, CIFAR-10, and Celeb-A, the ratio shown is with respect to a teacher generator size of depth $d=256$ (47,324,929 parameters, \emph{IS} = 7.02), $d=64$ (3,573,697 parameters, \emph{FID} = 7.42), and $d=128$ (12,652,417 parameters, \emph{FID} = 4.39) respectively. All teacher GANs were trained with a discriminator of corresponding teacher depth.}
    \centering
    \begin{tabular}{ c l r r r r r r r r} 
        \toprule
        & & \multicolumn{8}{c}{GAN Size ($d$)} \\
        & & 2 &	4 &	8 &	16 & 32 & 48 & 64 & 128 \\
        \hline
        \multicolumn{2}{c}{\multirow{2}{*}{\centering\shortstack{No. of Parameters}}} & 
            \multirow{2}{*}{28,351} & \multirow{2}{*}{62,077} & \multirow{2}{*}{145,657} &
            \multirow{2}{*}{377,329} & \multirow{2}{*}{109,8721} & \multirow{2}{*}{216,4177} & \multirow{2}{*}{3,573,697} & \multirow{2}{*}{12,652,417} \\ \\
        \hline
        \multirow{3}{*}{MNIST}
            & Ratio & 1669:1 & 762:1 & 325:1 & 125:1 & 43:1 & --- & 13:1 & 4:1 \\
            & IS (Stu.) & 5.80 & 6.41 & 6.60 & 6.83 & 6.87	& --- & 6.93 & 6.97 \\
            & IS (Reg.) & 1.86 & 3.63 & 4.73 & 5.07 & 6.08 & --- & 6.51 & 6.63 \\
        \hline
        \multirow{3}{*}{CIFAR-10} 
            & Ratio & 126:1 & 58:1 & 25:1 & 9:1 & 3:1 & 2:1 & --- & --- \\
            & FID (Stu.) & 11.76 & 11.00 & 9.57 & 8.39 & 7.80 & 7.58 & --- & --- \\
            & FID (Reg.) & 38.72 & 14.28 & 11.85 & 9.90 & 7.86 & 7.64 & --- & --- \\
        \hline
        \multirow{3}{*}{Celeb-A}
            & Ratio & 446:1 & 204:1 & 87:1 & 34:1 & 12:1 & 6:1 & 4:1 & ---\\
            & FID (Stu.) & 12.15 & 10.97 & 8.78 & 6.29 & 4.84 & --- & 4.54 & --- \\
            & FID (Reg.) & 45.49 & 18.72 & 11.06 & 9.14 & 5.05 & --- & 4.62 & --- \\
        \bottomrule
    \end{tabular}
    \label{table:compress_ratio}
\end{table*}


\emph{Qualitative Results}. Because of the deficiencies of the Inception Score and Frechet Inception Distance, it is important to qualitatively review the results. In Figures \ref{fig:output_comparison_mnist}, \ref{fig:output_comparison_cifar}, and \ref{fig:output_comparison_celeba}, we are able to see a direct comparison between the teacher, student, and a regular GAN of comparable size to the student. A visual review shows that the student is able to approximate the teacher without compromising the visual integrity of the image despite having a high compression ratio. One drawback however, is the presence of blur in the outputs from the student generator. This blur becomes more prominent as the compression ratio increases. The student generator is able to generate the basic structure of the image but fails to add details. This is unlike the control generator of the same size, which, although generates sharp images, misses on the basic image structure.


The student GAN thus outperforms the control GAN of the same size. This demonstrates the superiority of compressing an over-parameterized GAN, rather than training a small sized generator using the adversarial framework. 
Figure \ref{fig:mnist_interpolation} shows the output of interpolating the student and the teacher between two input vectors. We can see that the student is able to generate comparable images to that of the teacher's for each interpolation delta. This demonstrates that the student is learning to approximate the teacher's generation function, and not memorizing specific trained outputs.
\subsection{Joint Loss Training}
\begin{figure}[!ht]
    \centering
    \vskip 0.2in
    \centerline{\includegraphics[clip, trim=2.5cm 0cm 2.5cm 0cm, width=\columnwidth]{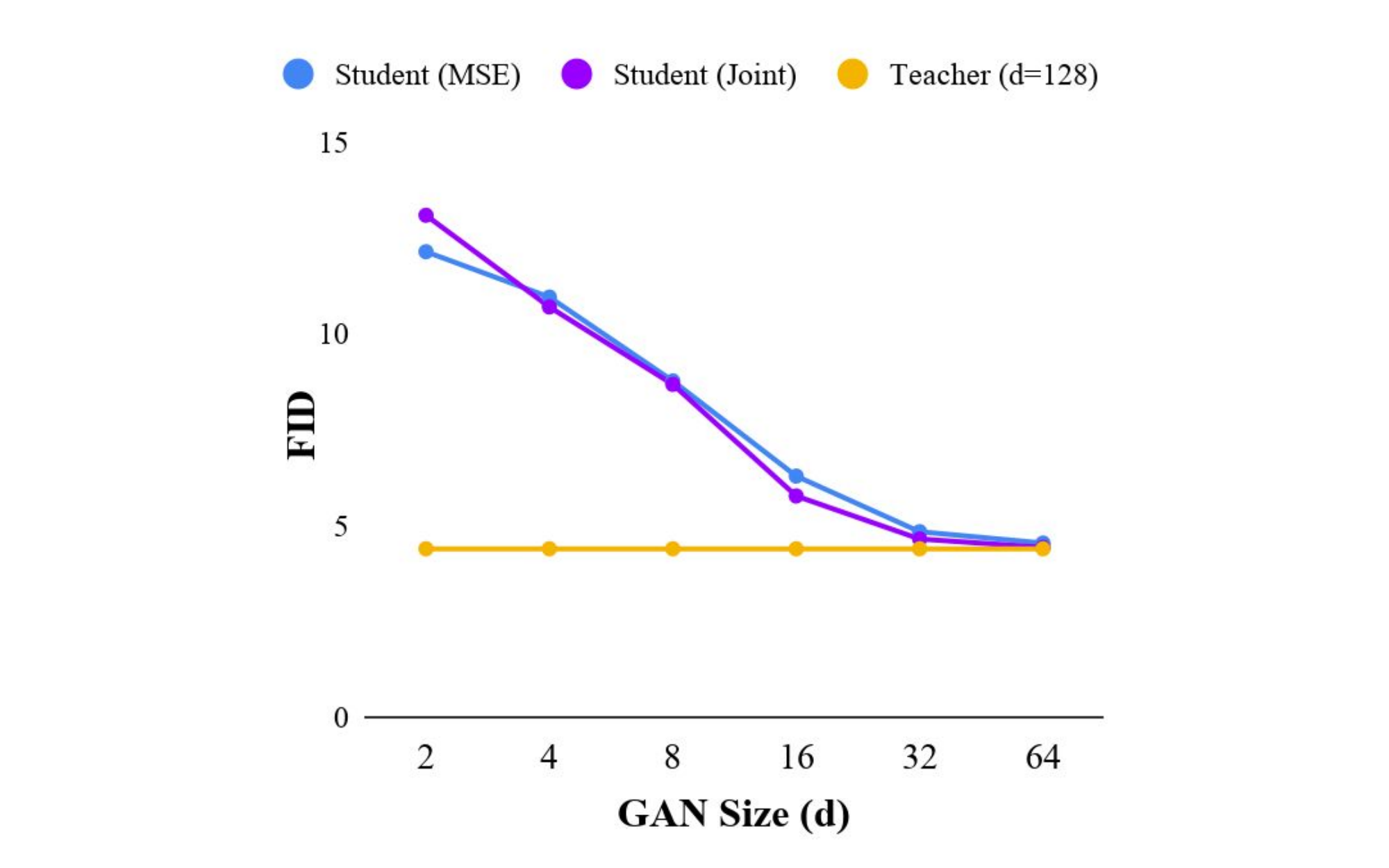}}
    \caption{Frechet Inception Distance comparison for various sizes of GANs using the joint training scheme. A low FID score is good.}
    \label{fig:fid_joint}
    \vskip -0.2in
\end{figure}

\begin{figure}[!h]
    \centering
    \vskip 0.2in
    \centerline{\includegraphics[clip, trim=2.5cm 0cm 2.5cm 0cm, width=\columnwidth]{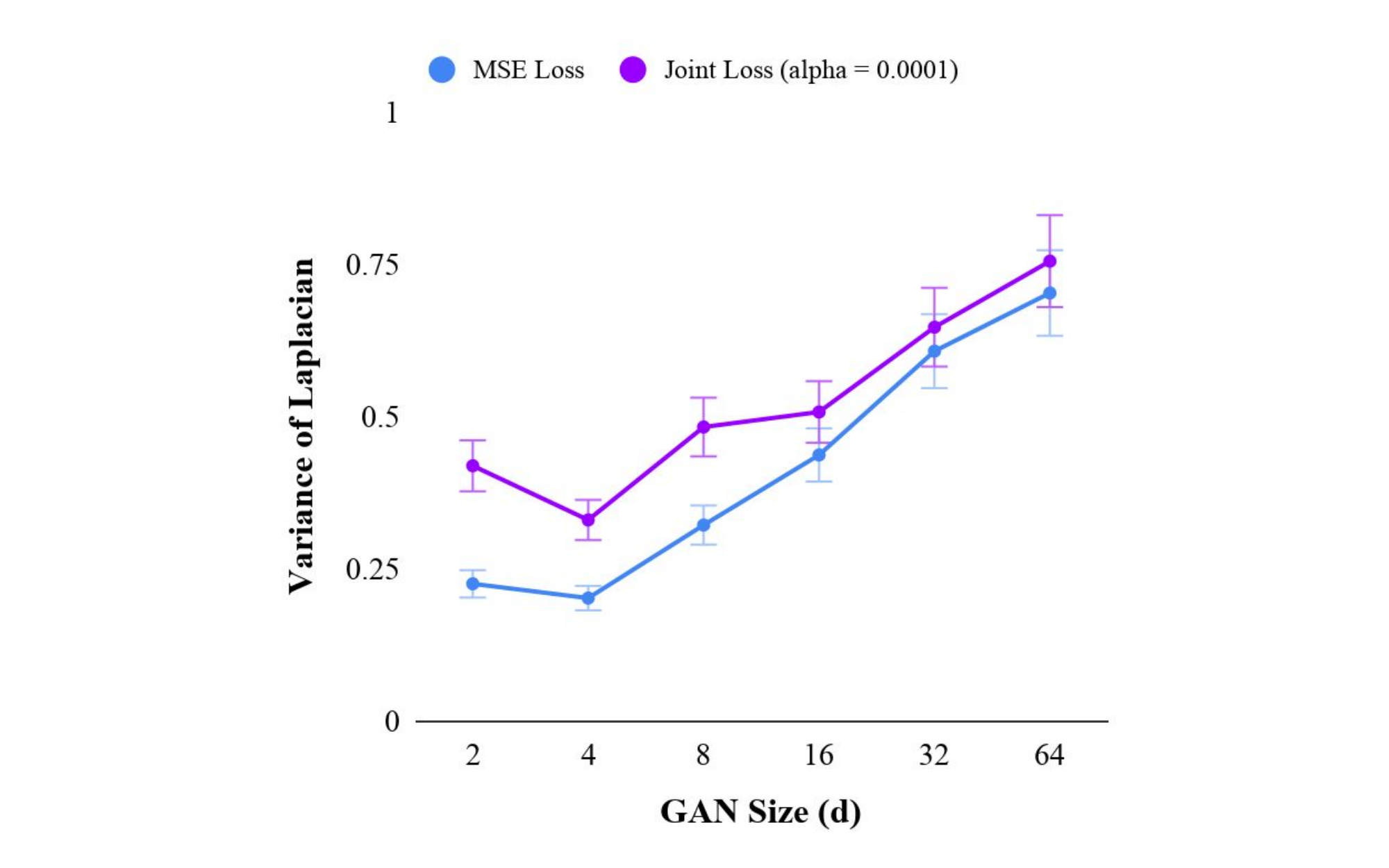}}
    \caption{Variance of Laplacian (VoL) comparison to evaluate blurriness in generated images from the student GAN trained using MSE loss, and the student GAN trained using joint loss. The values are reported as a ratio of the VoL score for student to the VoL score of the teacher, for both the students. The scores have been averaged across 10 different generated images. A higher value means the image is more sharp.}
    \label{fig:joint_vol}
    \vskip -0.2in
\end{figure}

\begin{figure}[!h]
    \centering
    \vskip 0.2in
    \centerline{\includegraphics[clip, trim=0cm 2cm 0cm 0cm, width=\columnwidth]{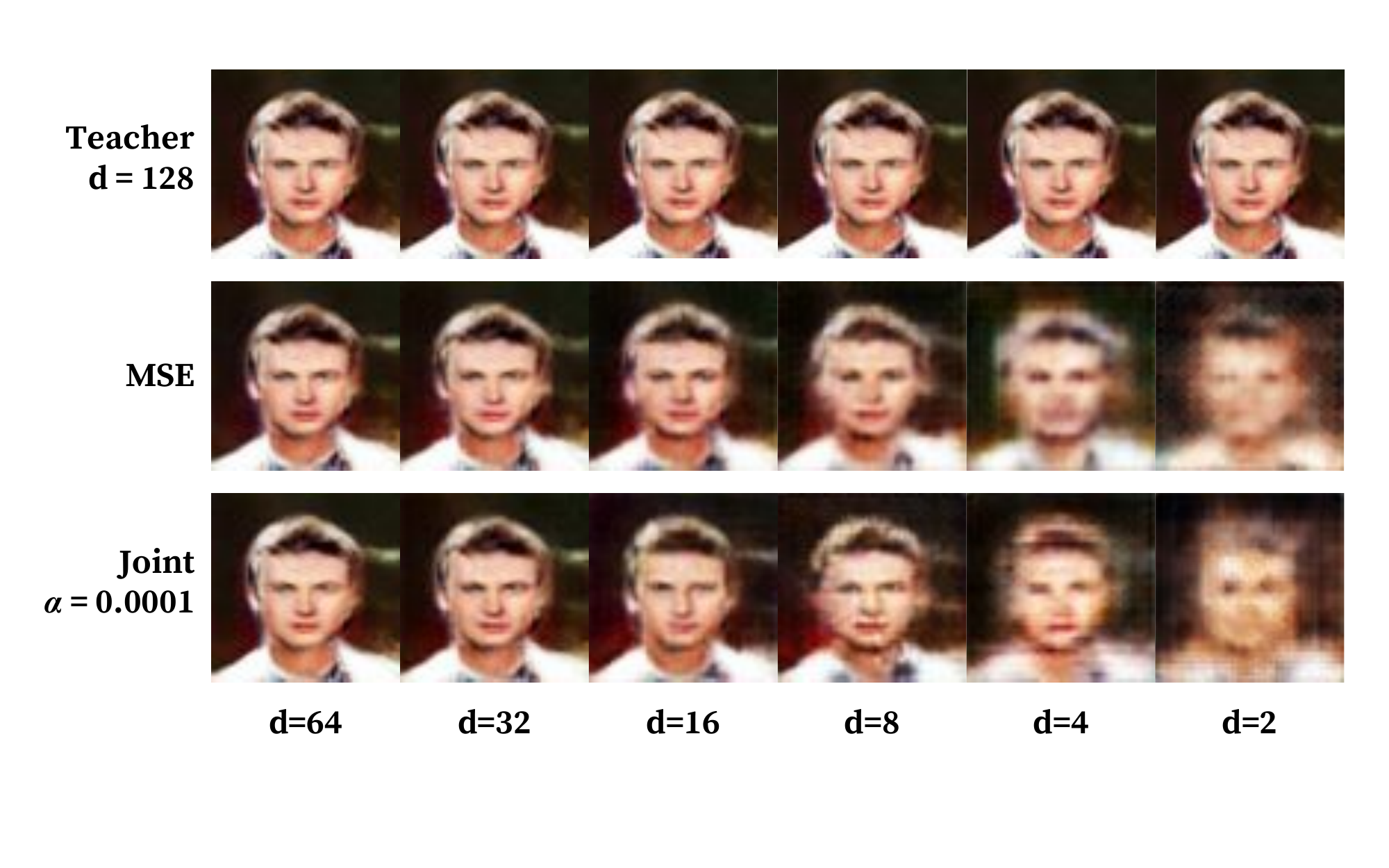}}
    \caption{Compression artifacts on the Celeb-A dataset from images generated from a teacher GAN ($d=128$), student GAN trained using MSE loss and student GAN trained using joint loss at $\alpha=0.0001$.}
    \label{fig:celeba_compression}
    \vskip -0.2in
\end{figure}

In Figure \ref{fig:fid_joint}, student GANs trained with joint loss outperform networks trained with only MSE loss in terms of slightly better FID scores. In Figure \ref{fig:joint_vol}, the joint loss GANs perform significantly better in terms of VoL metric, meaning that they produce much sharper images compared to the MSE loss GANs. This can also be observed visually through comparisons of the generated images in Figure \ref{fig:celeba_compression}.

\section{Discussion}

\subsection{Limit to Compression}
Through visual examination, the degradation of generated images seem to happen at higher compression ratio for more complex datasets, assuming that complexity grows in the following order: MNIST, CIFAR-10, Celeb-A. The bottom row of Figure \ref{fig:combined_quality_metrics_plot} show the outputs as a result of different compression ratios. The compressed MNIST GANs seem to have minimal compression artifacts across all compression levels whereas the compressed CIFAR-10 GANs seem to suffer significant degradation when $d=2$. For the compressed Celeb-A GANs, the degradation happens at an even lower compression ratio for Celeb-A GANs at $d=8$. The compression ratios referenced in the abstract is based on the smallest compressed GANs before significant observable degradation is present in the generated images. This observation suggests a potential limit to compression depending on the complexity of the data set. 
\subsection{Our Contributions}

Our work contributes to the topic of GAN compression. 
To summarize, we have made the following contributions in this paper:
\begin{itemize}
    \item We have developed two compression schemes for GANs using a student-teacher learning architecture (Figures \ref{fig:st_mse_loss}, \ref{fig:st_joint_loss}).
    
    \item We have evaluated the proposed compression methods over MNIST, CIFAR-10, and Celeb-A datasets. Our results show that the quality of generated imagery is maintained at high compression rates (1669:1, 58:1, 87:1 respectively) as measured by the Inception Score and Frechet Inception Distance metrics.
    
    \item We show that training a GAN of the same size without knowledge distillation produces comparatively diminished results, supporting the conjecture that over-parameterization is both helpful and necessary for neural networks to find a good function for GANs.  
    
    \item We observe a qualitative limit to GAN's compression for all the aforementioned datasets. We conjecture that there exists a fundamental compression limit of GANs similar to Shannon's compression theory \cite{MacKay2002InfoTheory}. 
    
\end{itemize}
\section{Conclusion}
Overall, we have demonstrated that applying the knowledge distillation method to GAN training can produce compressed generators without loss of quality or generalization. More specifically, we demonstrated that the student generators are able to outperform a traditionally trained GAN of the same size and approximate the underlying function of the teacher generator for the whole latent space. This further supports the necessity for over-parameterization when training an effective generator prior to distillation. Further, a qualitative limit to GAN compression has been observed for MNIST, CIFAR-10 and Celeb-A datasets.
\bibliography{main}
\bibliographystyle{icml2019}

\end{document}